\documentclass[runningheads]{llncs}
\usepackage[T1]{fontenc}
\usepackage{graphicx,verbatim}
\usepackage{amssymb}
\usepackage{caption}
\usepackage{booktabs}
\usepackage{bbold}
\usepackage{gensymb}
\usepackage{multirow}
\usepackage{multicol}
\usepackage{marvosym}
\usepackage[colorlinks,citecolor=black,linkcolor=blue]{hyperref}
\usepackage{color}

\begin{document}
\title{Contrastive Learning with Diffusion Features for Weakly Supervised Medical Image Segmentation}

\titlerunning{Contrastive Learning with Diffusion Features}

%

\author{
Dewen Zeng\inst{1}$^{(\textrm{\Letter})}$ \and
Xinrong Hu\inst{1} \and
Yu-Jen Chen\inst{2} \and
Yawen Wu\inst{1} \and
Xiaowei Xu\inst{3} \and
Yiyu Shi\inst{1}$^{(\textrm{\Letter})}$
}  
\authorrunning{D Zeng, et al.}

\institute{University of Notre Dame, Notre Dame, IN, USA \\\email{\{dzeng2, yshi4\}@nd.edu}\and
National Tsing Hua University, Taiwan\and
Guangdong Provincial People's Hospital, Guangzhou, China
}

\maketitle              
\begin{abstract}
Weakly supervised semantic segmentation (WSSS) methods using class labels often rely on class activation maps (CAMs) to localize objects. However, traditional CAM-based methods struggle with partial activations and imprecise object boundaries due to optimization discrepancies between classification and segmentation. Recently, the conditional diffusion model (CDM) has been used as an alternative for generating segmentation masks in WSSS, leveraging its strong image generation capabilities tailored to specific class distributions. By modifying or perturbing the condition during diffusion sampling, the related objects can be highlighted in the generated images. Yet, the saliency maps generated by CDMs are prone to noise from background alterations during reverse diffusion. To alleviate the problem, we introduce Contrastive Learning with Diffusion Features (CLDF), a novel method that uses contrastive learning to train a pixel decoder to map the diffusion features from a frozen CDM to a low-dimensional embedding space for segmentation. Specifically, we integrate gradient maps generated from CDM’s external classifier with CAMs to identify foreground and background pixels with fewer false positives/negatives for contrastive learning, enabling robust pixel embedding learning. Experimental results on four segmentation tasks from two public medical datasets demonstrate that our method significantly outperforms existing baselines.


\end{abstract}
\section{Introduction}


To reduce the labeling effort required for training fully supervised segmentation models, researchers have extensively explored Weakly Supervised Semantic Segmentation (WSSS) methods, which rely on more easily obtainable labels such as image-level labels \cite{wang2020self,li2021pseudo,jiang2021layercam,xie2022c2am,chen2023ame}. 
A prevalent strategy within WSSS involves generating pixel-level segmentations from image-level labels using Class Activation Maps (CAMs) \cite{zhou2016learning} or its variants \cite{selvaraju2017grad,chattopadhay2018grad,omeiza2019smooth,wang2020score,jiang2021layercam,chen2023ame}. 
However, CAM-based approaches suffer from under-activation and imprecise object boundary problems due to the gap between full and weak supervision \cite{wang2020self,yoon2024diffusion}. These challenges are even more pronounced in medical imaging, where images tend to be lower in resolution and objects are often small. To improve CAM's localization performance for small objects, LayerCAM \cite{jiang2021layercam} combines activation maps from multiple convolution layers to generate the final CAM. Similarly, AME-CAM \cite{chen2023ame} uses a multi-exit training strategy to extract activation maps from different layers and aggregate them using pixel-wise weights learned by contrastive learning. Despite these advancements, the issue of imprecise object boundaries remains unresolved.

\begin{figure}[t!]
	\centering
	\includegraphics[width=0.95\linewidth]{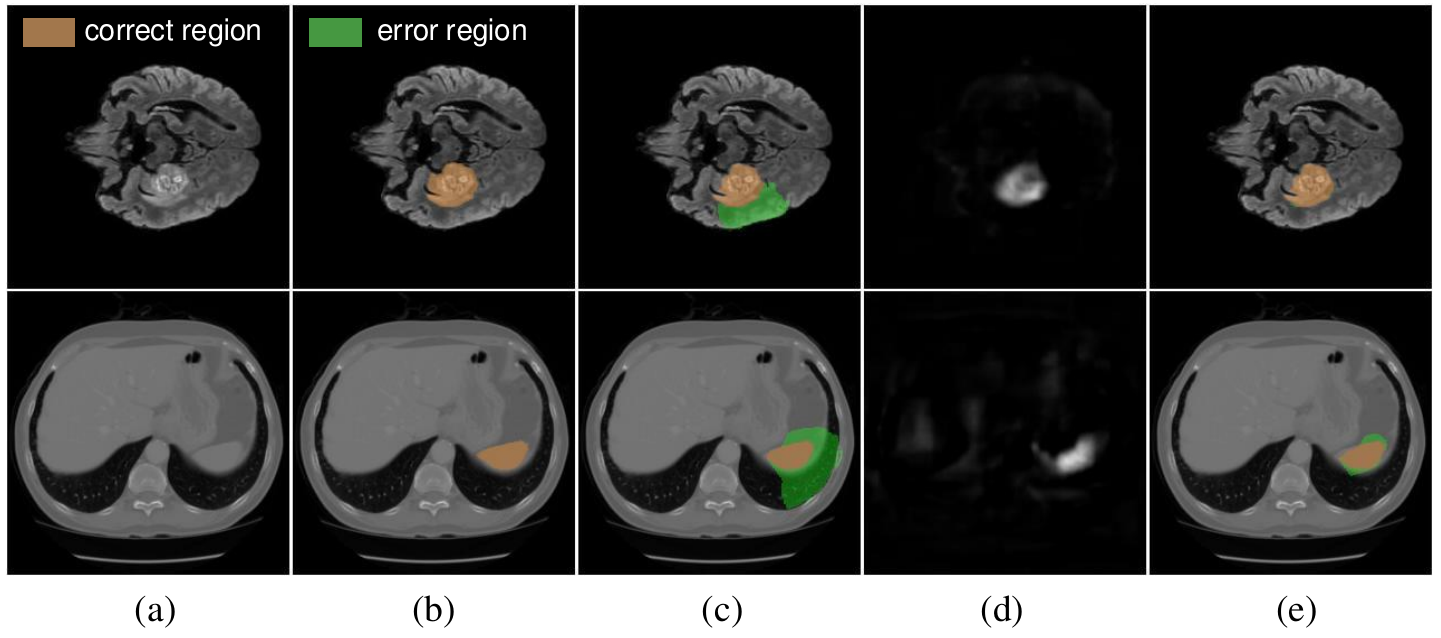}
	\caption{(a) Original images (b) Ground truths (c) Segmentations generated by CAM (d) Gradient maps generated by the external classifier of the conditional diffusion model (e) Refined segmentations by fusing CAM and the gradient map. Images are from the BraTS21 \cite{baid2021rsna} and FLARE21 \cite{ma2022fast} datasets.}
	\label{fig:motivation}
\end{figure}

Meanwhile, Denoising Diffusion Probabilistic Models (DDPM) have demonstrated superior image generation capabilities \cite{ho2020denoising,dhariwal2021diffusion,ho2022classifier,rombach2022high,nichol2021improved}. Several studies have explored using DDPM for WSSS with only image-level labels \cite{wolleb2022diffusion,hu2023conditional,yoon2024diffusion}.
\cite{wolleb2022diffusion} utilizes an external classifier to guide DDPMs to remove the anomalous regions so that the anomalies can be identified by comparing the reconstructed and original images. 
\cite{hu2023conditional} perturbs the condition embedding in the conditional diffusion model (CDM) during reverse diffusion to highlight the related object in the gradient map.
Compared to CAM, DDPM-based object localization could be more precise, as the generation process tends to focus on class-related areas when class conditioning is enforced. However, due to the nature of diffusion sampling, the segmentation masks produced by DDPMs are susceptible to background noise. 
DiG \cite{yoon2024diffusion} refines CAM generated by Vision Transformer (ViT) using features extracted from DDPMs, enhancing its localization performance. However, since their CAMs are derived from the final classification layer at a low resolution, they are less suitable for segmenting small objects in medical imaging.

In this work, we propose a novel Contrastive Learning with Diffusion Features (CLDF) framework to generate better segmentation masks for medical image WSSS tasks using only image-level labels. Similar to \cite{yoon2024diffusion}, we first extract raw features from a frozen CDM, as many works have shown that these features contain high-level semantic features suitable for segmentation \cite{baranchuk2021label,graikos2022diffusion,pinaya2022fast,wolleb2022diffusion}. Instead of using these features to refine the CAM generated by CNN or ViT, we aim to learn a pixel decoder that maps the extracted pixel features to a lower-dimension embedding space for segmentation purposes. To ensure the pixel decoder learns discriminative features for foreground and background pixels, we apply a contrastive loss to pixels selected by fusing CAM and the gradient map generated by an external classifier from CDM. CLDF is motivated by our observation that CAM effectively localizes objects but lacks precise boundaries, whereas the gradient map provides sharper object boundaries but may highlight unrelated background regions. By combining these two sources, we can mitigate their individual limitations and produce a more reliable activation region (see Fig. \ref{fig:motivation}). Once the pixel decoder is trained, K-means clustering is applied to the embeddings to generate the final segmentation mask. In line with previous WSSS works \cite{chen2023ame,hu2023conditional}, we focus on binary segmentation tasks, where the target object is either present or absent in the image. Since our approach relies on the quality of the CAMs and can be integrated with any CAM method to generate segmentation masks, it falls under the \textbf{CAM refinement} (post-processing) phase rather than CAM improvement, which focuses on generating better CAMs.

\section{Contrastive Learning with Diffusion Features}


\begin{figure}[t!]
	\centering
	\includegraphics[width=0.95\linewidth]{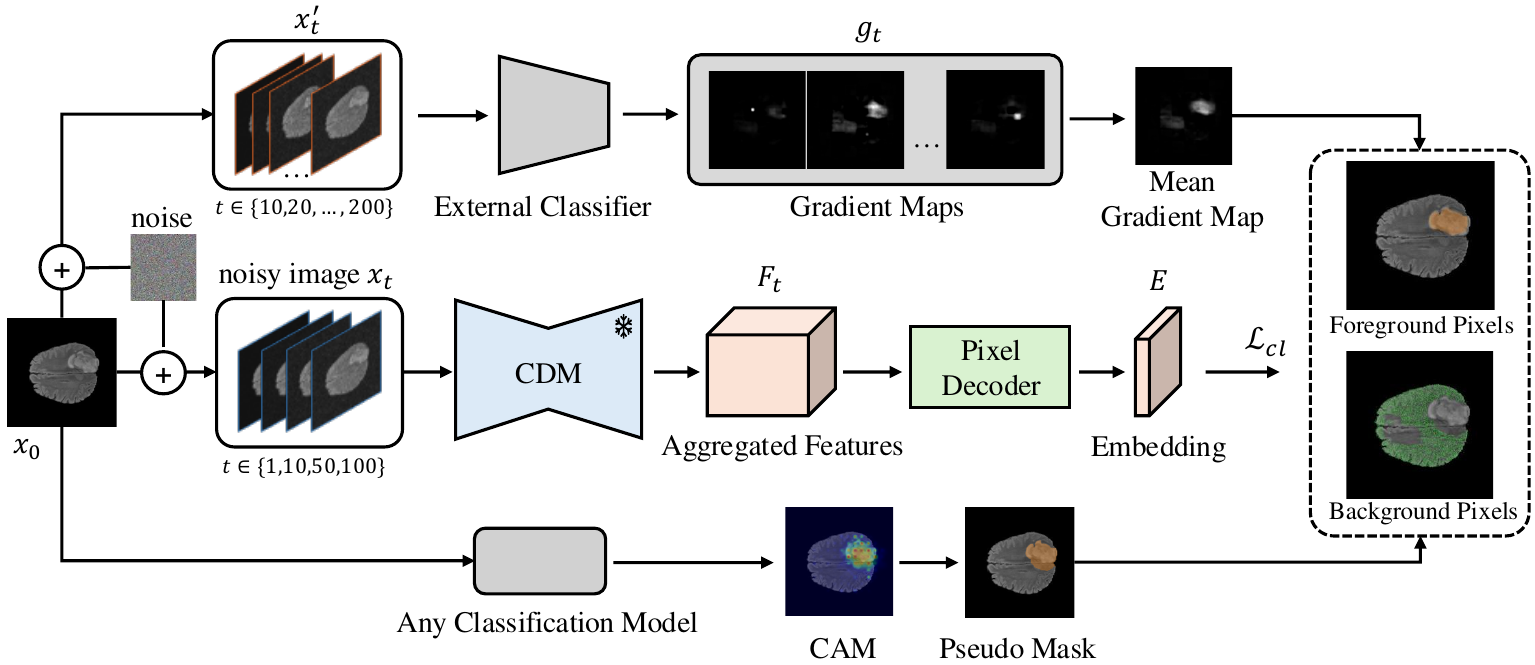}
	\caption{An overview of the proposed CLDF method. The pixel decoder is trained with a contrastive loss $\mathcal{L}_{cl}$ to map the aggregated features $F_{t}$ extracted from a Conditional diffusion model (CDM) to a lower-dimension embedding space $E$. The contrastive loss utilizes foreground and background pixel features selected by fusing the CAM generated by any classification model and the mean gradient map generated by the external classifier of the CDM.}
	\label{fig:overview}
\end{figure}

Inspired by the observation in \cite{baranchuk2021label} that a well-trained pixel classifier can effectively decode the diffusion features for semantic segmentation, we show that such a pixel classifier/decoder can be learned even without ground truth masks by leveraging contrastive learning on pixel features selected from CAMs and gradient maps generated from classification models.

As shown in Fig. \ref{fig:overview}, given a medical image $x_{0} \in \mathbb{R}^{H\times W\times C}$, we first add random noise at different timesteps $t$ to get noisy images $x_{t}$, $t \in \{1,10,50,100\}$. The noisy images are then used to extract features from various layers of the decoder of a pre-trained CDM. The extracted features are unsampled to $H\times W$ and concatenated to form $F_{t} \in \mathbb{R}^{H\times W\times D}$, where $D$ is the feature dimension for each pixel. Then, the features $F_t$ are averaged across different timesteps $t$ and passed through a pixel decoder to generate embeddings $E \in \mathbb{R}^{H\times W\times O}$, where $O$ is the final embedding dimension of each pixel. To train the pixel decoder, we employ contrastive learning on the foreground and background pixel features selected from $E$. Specifically, the foreground pixels are selected by the intersection of activated regions from CAM and the mean gradient map. The CAM is obtained using a classification model with any existing CAM method, while the gradient map $g_{t}$ is calculated by deriving the gradient towards the positive class with respect to the input image $x_t^{\prime}$ at timestep $t$ in the external classifier of CDM. Similar to \cite{dhariwal2021diffusion}, the gradient map highlights the relevant object within the noisy image to guide diffusion sampling. To reduce noise in the gradient map from a single timestep, we take the mean gradient map across multiple timesteps $t \in \{10,20,...,200\}$. Intuitively, CAM effectively localizes the object but lacks precise boundaries, while the mean gradient map provides sharper object boundaries but introduces background noise. By combining the strengths of both methods, we can obtain a refined activation map that enhances localization while maintaining high confidence in the identified regions.
Meanwhile, the background pixels are selected from the regions not activated by either CAM or the mean gradient map. To optimize memory usage during training, we randomly sample 5,000 background pixels for contrastive learning. 
By focusing on high-confidence pixel embeddings, we mitigate the adverse effects of false positives/negatives, leading to improved segmentation accuracy. Given that the aggregated features $F_{t}$ are highly representative, the pixel decoder can effectively adapt to uncertain regions. 
We use supervised contrastive loss \cite{khosla2020supervised} to train the pixel decoder:

\begin{equation}
    \mathcal{L}_{cl}=\sum_{i=1}^{N}-\frac{1}{|\Omega_{i}^{+}|}\sum_{j\in \Omega_{i}^{+}}{log\frac{e^{sim(z_i,z_j)/\tau}}{\sum_{k=1}^{N}\mathbb{1}_{i\neq k}\cdot e^{sim(z_i,z_k)/\tau}}}.
\label{eq:1}
\end{equation}

where $N$ is the number of pixels selected for contrastive learning in the current batch, which may vary during training. $z$ denotes the pixel embedding selected from $E$. $\Omega_{i}^{+}$ is the set of indices of positive samples corresponding to ${z}_i$. $\tau$ is a temperature scaling parameter. 

During inference, the pre-trained CDM and pixel decoder are used to generate pixel embeddings. Then, K-means clustering is applied to these embeddings to produce binary segmentation masks. It is worth noting that our CLDF method is flexible and can be integrated with any existing technique as a post-processing approach to improve the accuracy of segmentation masks.

\section{Experiments and Discussions}

\subsection{Datasets and Implementation Details}

\textbf{BraTS21 Dataset.} 
We use the Brain Tumor Segmentation Challenge (BraTS) Task1 in 2021 \cite{baid2021rsna} to evaluate the effectiveness of our method. The dataset comprises 2,000 cases of brain scans, with each case containing four 3D volumes from four different imaging modalities. We use the same pre-processing setting as in \cite{chen2023ame,hu2023conditional}, resulting in a total of 193,905 slices in the training and evaluation set and 6,875 slices in the testing set (5,802 positive and 1,073 negative). All tumor types are treated as a single class to frame the problem as a binary segmentation task. Unlike \cite{chen2023ame,hu2023conditional}, which used slices with a single modality, we concatenate slices from all four modalities into a 4-channel image to serve as the input for our model, simulating how radiologists evaluate MRI scans in real-world scenarios.

\textbf{FLARE21 Dataset.} The MICCAI 2021 Fast and low GPU memory abdominal organ segmentation (FLARE) dataset \cite{ma2022fast} contains 361 cases from 11 medical centers, each of which includes a 3D CT scan with segmentation labels for the kidney, spleen, liver, and pancreas. In this work, we focus on the binary segmentation task for the kidney, spleen, and liver. For each task, only the target structure is treated as the foreground, while all other structures are treated as the background, even if they are present. We randomly partitioned the official training set into 211, 50, and 100 for training, validation, and testing.

\textbf{Implementation Details.} We adopt the same setup as described in \cite{hu2023conditional,dhariwal2021diffusion} to train the UNet-based CDM and the external classifier. The external classifier uses the same backbone as the CDM encoder but operates without conditioning. We train the diffusion model for four days with a batch size of 8 and the external classifier for 100,000 iterations with a batch size of 16. The AdamW optimizer is used with a learning rate of 1e-4 and 3e-4 to train the CDM and the external classifier, respectively. The pixel decoder consists of a 4-layer multilayer perception (MLP) with a structure of ($D$,16,16,16,16), where $D$ is the feature dimension for each pixel in $F_t$. We train the pixel decoder using an SGD optimizer with a learning rate of $1.0$ for 5 epochs, and the batch size is set to 4. The temperature parameter $\tau$ in Equation \ref{eq:1} is 0.1. We use AME-CAMs \cite{chen2023ame} for Brats21 and LayerCAM \cite{jiang2021layercam} for FLARE21 to select pixels for CLDF as they empirically perform better. Following the evaluation protocol in \cite{chen2023ame,hu2023conditional}, dice score and Intersection over Union (IoU) are used as segmentation evaluation metrics. All 2D images are resized and center-cropped to 256$\times$256 for consistency. Only image-level classification labels are utilized during the training and evaluation. The ground truth segmentation masks are used in the testing stage.

\subsection{Comparison with State-of-the-art}

\begin{table}[t]
\setlength\tabcolsep{4.0pt}
\centering
\caption{Comparisons with state-of-the-art WSSS methods on BraTS21 and FLARE21 Datasets. Results are reported in the form of mean$\pm$std. FSL means fully supervised learning.}

\begin{tabular}{l|cc|cc}
\toprule
& \multicolumn{2}{c}{BraTS21} & \multicolumn{2}{c}{FLARE21 Kidney} \\ 
Method & Dice $\uparrow$ & IoU $\uparrow$ & Dice $\uparrow$ & IoU $\uparrow$ \\ \hline
GradCAM (2017) \cite{selvaraju2017grad} & 0.392$\pm$0.29 & 0.286$\pm$0.24 & 0.335$\pm$0.31 & 0.251$\pm$0.27 \\
ScoreCAM (2020) \cite{wang2020score} & 0.318$\pm$0.11 & 0.195$\pm$0.08 & 0.208$\pm$0.13 & 0.122$\pm$0.08 \\
SEAM (2021) \cite{wang2020self} & 0.342$\pm$0.08 & 0.210$\pm$0.06 & 0.257$\pm$0.09 & 0.123$\pm$0.06 \\
LayerCAM (2021) \cite{jiang2021layercam} & 0.670$\pm$0.15 & 0.521$\pm$0.16 & 0.514$\pm$0.23 & 0.380$\pm$0.22 \\
CDM (2023) \cite{hu2023conditional} & 0.563$\pm$0.02 & 0.450$\pm$0.02 & 0.322$\pm$0.19 & 0.207$\pm$0.14 \\
AME-CAM (2023) \cite{chen2023ame} & 0.827$\pm$0.14 & 0.725$\pm$0.18 & 0.413$\pm$0.17 & 0.288$\pm$0.13  \\
DiG (2024) \cite{yoon2024diffusion} & 0.535$\pm$0.17 & 0.381$\pm$0.15 & 0.378$\pm$0.14 & 0.242$\pm$0.11 \\ \hline
CLDF (ours) & \textbf{0.880$\pm$0.11} & \textbf{0.798$\pm$0.14} & \textbf{0.740$\pm$0.20} & \textbf{0.621$\pm$0.21} \\ \hline
FSL (UNet) & 0.920$\pm$0.09 & 0.861$\pm$0.12 & 0.949$\pm$0.08 & 0.910$\pm$0.09 \\
\bottomrule
\end{tabular}

\begin{tabular}{l|cc|cc}
\toprule
& \multicolumn{2}{c}{FLARE21 Spleen} & \multicolumn{2}{c}{FLARE21 Liver} \\ 
Method & Dice $\uparrow$ & IoU $\uparrow$ & Dice $\uparrow$ & IoU $\uparrow$ \\ \hline
GradCAM (2017) \cite{selvaraju2017grad} & 0.243$\pm$0.29 & 0.175$\pm$0.23 & 0.261$\pm$0.22 & 0.170$\pm$0.16 \\
ScoreCAM (2020) \cite{wang2020score} & 0.138$\pm$0.13 & 0.080$\pm$0.08 & 0.438$\pm$0.19 & 0.299$\pm$0.16 \\
SEAM (2020) \cite{wang2020self} & 0.114$\pm$0.09 & 0.063$\pm$0.05 & 0.408$\pm$0.18 & 0.271$\pm$0.14 \\
LayerCAM (2021) \cite{jiang2021layercam} & 0.492$\pm$0.21 & 0.352$\pm$0.19 & 0.721$\pm$0.17 & 0.587$\pm$0.18 \\
CDM (2023) \cite{hu2023conditional} & 0.374$\pm$0.20 & 0.249$\pm$0.16 & 0.392$\pm$0.19 & 0.260$\pm$0.14  \\
AME-CAM (2023) \cite{chen2023ame} & 0.560$\pm$0.13 & 0.399$\pm$0.12 & 0.556$\pm$0.18 & 0.406$\pm$0.17 \\
DiG (2024) \cite{yoon2024diffusion} & 0.313$\pm$0.23 & 0.208$\pm$0.17 & 0.470$\pm$0.19 & 0.326$\pm$0.16 \\ \hline
CLDF (ours) & \textbf{0.623$\pm$0.15} & \textbf{0.470$\pm$0.16} & \textbf{0.778$\pm$0.15} & \textbf{0.658$\pm$0.18} \\ \hline
FSL (UNet) & 0.936$\pm$0.16 & 0.905$\pm$0.17 & 0.952$\pm$0.12 & 0.923$\pm$0.13 \\
\bottomrule
\end{tabular}

\label{table:sota}
\end{table}

\begin{figure}[t!]
	\centering
	\includegraphics[width=1.0\linewidth]{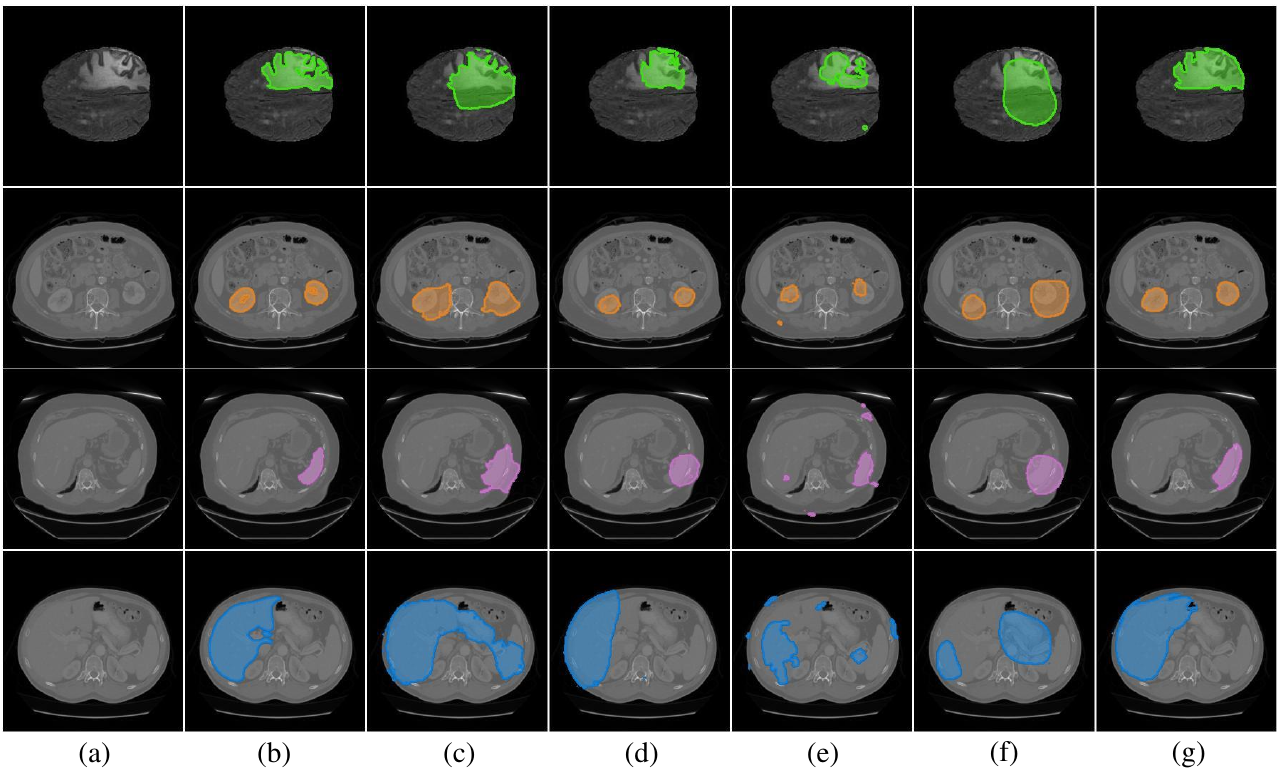}
	\caption{Visualization of segmentation results. (a) Original images (b) Ground truths (c) LayerCAM \cite{jiang2021layercam} (d) AME-CAM \cite{chen2023ame} (e) CDM \cite{hu2023conditional} (f) DiG \cite{yoon2024diffusion} (g) CLDF (ours). From top to bottom, images are from BraTS21 \cite{baid2021rsna}, FLARE21 \cite{ma2022fast} Kidney, Spleen, and Liver segmentation tasks, respectively.}
	\label{fig:qualitative_analysis}
\end{figure}

In this section, we compare the proposed CLDF method with seven state-of-the-art WSSS methods, namely GradCAM \cite{selvaraju2017grad}, ScoreCAM \cite{wang2020score}, LayerCAM \cite{jiang2021layercam}, SEAM \cite{wang2020self}, CDM \cite{hu2023conditional}, AME-CAM \cite{chen2023ame} and DiG \cite{yoon2024diffusion}. 
Fully supervised learning results are also reported as the upper bound of all WSSS methods.

From Table \ref{table:sota}, we can observe that (1) CAMs extracted from the last convolutional layer (e.g., ScoreCAM, SEAM, and DiG) generally produce lower Dice scores across all tasks except FLARE21 Liver, where the liver is relatively large. This suggests that the low-resolution CAMs are less suitable for WSSS problems with small objects. (2) Methods that aggregate CAMs from multiple convolutional layers (e.g., LayerCAM, AME-CAM) improve Dice scores, meaning that mid-layer features contribute to better object localization. (3) Our proposed CLDF consistently achieves optimal results across all tasks. (4) A performance gap remains between WSSS methods and fully supervised learning (FSL). The gap is larger in challenging tasks like FLARE21 spleen segmentation, indicating that there is still plenty of room for improvement for WSSS methods. 
Fig. \ref{fig:qualitative_analysis} presents qualitative segmentation comparisons among five WSSS methods. It can be seen that CDM and DiG are prone to have false positives from unrelated regions. While LayerCAM and AME-CAM effectively localize target objects, their object boundaries remain imprecise. In contrast, our CLDF approach reduces the under-estimated regions and provides more accurate object boundaries.

\subsection{Ablation Study}

\begin{table}[t!]
\setlength\tabcolsep{4.0pt}
\centering
\caption{Comparisons of features extracted from different pre-trained models. UNet encoder is the external classifier of the CDM.}

\begin{tabular}{l|cc|cc}
\toprule
& \multicolumn{2}{c}{BraTS21} & \multicolumn{2}{c}{FLARE21 Kidney} \\ 
Pre-trained Model & Dice $\uparrow$ & IoU $\uparrow$ & Dice $\uparrow$ & IoU $\uparrow$ \\ \hline
ResNet18 & 0.640$\pm$0.32 & 0.542$\pm$0.30 & 0.448$\pm$0.25 & 0.320$\pm$0.20  \\
ResNet50 & 0.775$\pm$0.15 & 0.652$\pm$0.16 & 0.644$\pm$0.20 & 0.504$\pm$0.20 \\
UNet Encoder & 0.760$\pm$0.13 & 0.629$\pm$0.15 & 0.605$\pm$0.21 & 0.465$\pm$0.21 \\
CDM & \textbf{0.880$\pm$0.11} & \textbf{0.798$\pm$0.14} & \textbf{0.740$\pm$0.20} & \textbf{0.621$\pm$0.21} \\
\bottomrule
\end{tabular}

\label{table:diffusion_features}
\end{table}

\subsubsection{The importance of diffusion features.}
To demonstrate the significance of diffusion features for WSSS, we compare segmentation results with features extracted from other pre-trained models, including ResNet18, ResNet50, and UNet Encoder (the external classifier of CDM), all pre-trained with only image-level labels. The results on the BraTS21 and FLARE21 Kidney datasets are summarized in Table \ref{table:diffusion_features}. It can be seen that features extracted from the diffusion model yield superior segmentation performance on both datasets. This aligns with the finding of \cite{baranchuk2021label}, which highlights that diffusion features inherently capture semantic information beneficial for segmentation tasks.

\begin{table}[t]
\setlength\tabcolsep{5.0pt}
\centering
\caption{Comparisons of using contrastive learning (CL) on pixels selected from different methods. MG refers to the mean gradient map.}

\begin{tabular}{l|cc|cc}
\toprule
& \multicolumn{2}{c}{BraTS21} & \multicolumn{2}{c}{FLARE21 Kidney} \\ 
Method & Dice $\uparrow$ & IoU $\uparrow$ & Dice $\uparrow$ & IoU $\uparrow$ \\ \hline
LayerCAM & 0.670$\pm$0.15 & 0.521$\pm$0.16 & 0.514$\pm$0.23 & 0.380$\pm$0.22  \\
AME-CAM & 0.827$\pm$0.14 & 0.725$\pm$0.18 & 0.413$\pm$0.17 & 0.288$\pm$0.13  \\
MG & 0.692$\pm$0.12 & 0.541$\pm$0.13 & 0.339$\pm$0.16 & 0.263$\pm$0.13 \\
LayerCAM+CL & 0.776$\pm$0.13 & 0.650$\pm$0.16 & 0.692$\pm$0.30 & 0.598$\pm$0.30 \\
AME-CAM+CL & 0.847$\pm$0.14 & 0.755$\pm$0.17 & 0.631$\pm$0.15 & 0.476$\pm$0.14 \\
MG+CL & 0.800$\pm$0.11 & 0.680$\pm$0.14 & 0.622$\pm$0.21 & 0.482$\pm$0.21 \\
LayerCAM+MG+CL & 0.795$\pm$0.12 & 0.675$\pm$0.15 & 
\textbf{0.740$\pm$0.20} & \textbf{0.621$\pm$0.21} \\
AME-CAM+MG+CL & \textbf{0.880$\pm$0.11} & \textbf{0.798$\pm$0.14} & 0.698$\pm$0.15 & 0.554$\pm$0.16 \\
\bottomrule
\end{tabular}

\label{table:mean_gradient}
\end{table}

\begin{figure}[t!]
	\centering
	\includegraphics[width=1.0\linewidth]{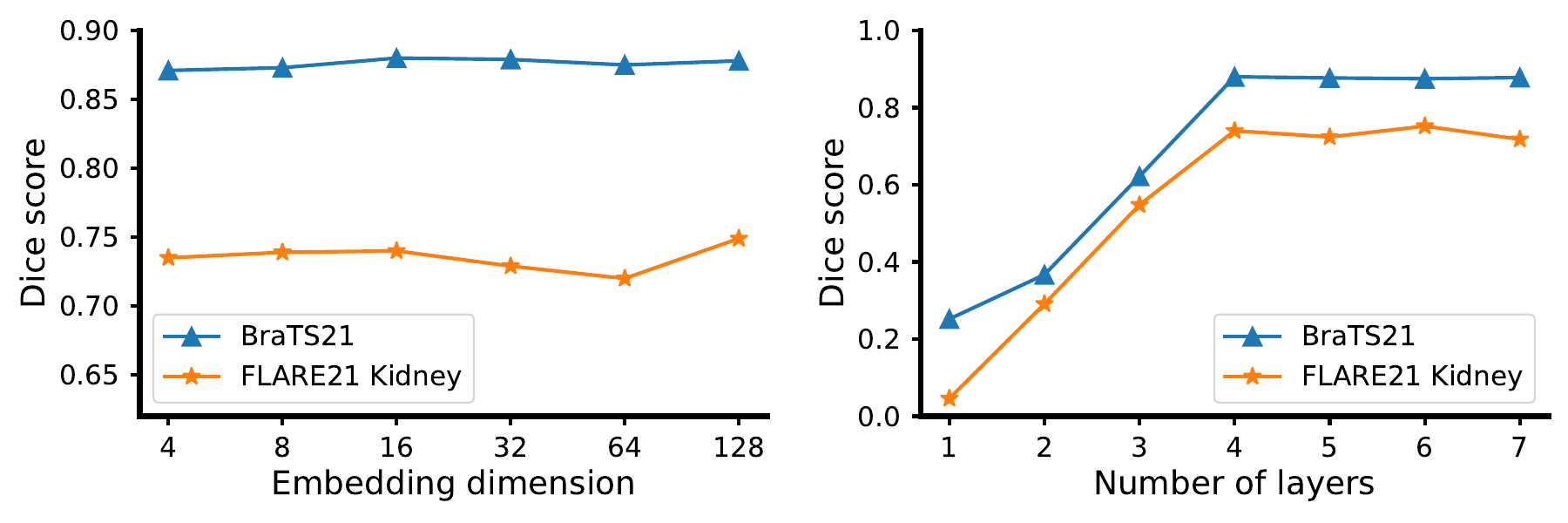}
	\caption{Ablation study on the pixel decoder structure: (Left) Impact of output embedding dimension on the Dice score, with the number of layers in MLP fixed to 4. (Right) Effect of the number of layers in MLP on the Dice score, with both the hidden and output dimensions fixed at 16.}
	\label{fig:ablation_study}
\end{figure}

\subsubsection{The importance of fusing CAM and the mean gradient map.}
In Table \ref{table:mean_gradient}, we evaluate the segmentation performance of the pixel decoder using different combinations of CAMs and mean gradients to identify pixel features. We can see that contrastive learning alone can improve the segmentation masks generated by either CAM or the mean gradient. Fusing them can significantly reduce false positives/negatives in the contrastive loss, thus resulting in better performance. 
Furthermore, higher-quality CAMs generally result in better CLDF results.

\subsubsection{Pixel decoder structure.} In Fig. \ref{fig:ablation_study}, we conduct ablation studies exploring variations in the pixel decoder structure. The results indicate that the output embedding dimension has minimal impact on the final Dice score, suggesting that a small dimension is sufficient to represent pixel features for segmentation. While a deeper pixel decoder does not guarantee improved performance, at least four layers are required to effectively decode the raw diffusion features.

\section{Conclusion}

In this paper, we propose a novel Contrastive Learning with Diffusion Features (CLDF) method for WSSS in medical imaging. By fusing CAM and the mean gradient to identify high-confidence foreground and background pixels, CLDF uses contrastive learning to train a pixel decoder to effectively map diffusion features to a lower-dimensional embedding space for better segmentation. Experiments on the BraTS21 and FLARE21 datasets demonstrate that our CLDF achieves state-of-the-art segmentation performance across various WSSS tasks. 

%
%
%
\bibliographystyle{splncs04}
\bibliography{reference}
%




\end{document}